\newcommand{\parshrinky}{\vspace{-5mm}}
\newcommand{\parskiny}{\vspace{-2mm}}
\newcommand{\seckiny}{\vspace{-2mm}}
\newcommand{\figskiny}{\vspace{-4mm}}

\newcommand{\changed}[1]{#1}

\newcommand{\done}[1]{}

\documentclass[10pt,twocolumn,letterpaper,english]{article}

\usepackage{subfigure}
\usepackage{cvpr}
\usepackage{times}
\usepackage{epsfig}
\usepackage{graphicx}
\usepackage{amsmath}
\usepackage{amssymb}
\usepackage{verbatim}
\usepackage{rotating}
\usepackage{float}
\usepackage{graphicx}
\usepackage{amsmath}
\usepackage{bm}

\DeclareMathOperator*{\argmin}{arg\,min}
\DeclareMathOperator*{\argmax}{arg\,max}

\usepackage[pagebackref=false,breaklinks=true,letterpaper=true,colorlinks,bookmarks=false]{hyperref}

\cvprfinalcopy 

\def\cvprPaperID{317} 

\ifcvprfinal\fi
\begin{document}

\title{Associative embeddings for large-scale knowledge transfer with self-assessment}

\author{ Alexander Vezhnevets \hspace{1cm} Vittorio Ferrari\\
The University of Edinburgh\\
Edinburgh, Scotland, UK\\
}

\providecommand{\tabularnewline}{\\}
\floatstyle{ruled}
\newfloat{algorithm}{tbp}{loa}
\floatname{algorithm}{Algorithm}



\maketitle

\begin{abstract}
\vspace{-3mm}
We propose a method for knowledge transfer between semantically related classes in ImageNet. By transferring knowledge from the images that have bounding-box annotations to the others, our method is capable of automatically populating ImageNet with many more bounding-boxes.The underlying assumption that objects from semantically related classes look alike is formalized in our novel Associative Embedding (AE) representation. AE recovers the latent low-dimensional space of appearance variations among image windows. The dimensions of AE space tend to correspond to aspects of window appearance (e.g. side view, close up, background). We model the overlap of a window with an object using Gaussian Processes (GP) regression, which spreads annotation smoothly through AE space. The probabilistic nature of GP allows our method to perform self-assessment, i.e. assigning a quality estimate to its own output. It enables trading off the amount of returned annotations for their quality. A large scale experiment on 219 classes and 0.5 million images demonstrates that our method outperforms state-of-the-art methods and baselines for object localization. Using self-assessment we can automatically return bounding-box annotations for 51\% of all images with high localization accuracy (i.e. 71\% average overlap with ground-truth).
\end{abstract}

\vspace{-6mm}

\section{Introduction}
\seckiny
The large, hierarchical ImageNet dataset~\cite{Deng:CVPR2009} presents new challenges and opportunities for computer vision. Although the dataset contains over 14 million images, only a fraction of them has bounding-box annotations ($10\%$) and none have segmentations (object outlines). Automatically populating ImageNet with bounding-boxes or segmentations is a  challenging problem, which has recently drawn attention~\cite{Guillaumin2014ijcv,Guillaumin2012cvpr}.
These annotations could be used as training data for problems such as object class detection~\cite{Dalal05:thomas}, tracking~\cite{leibe07iccv} and pose estimation~\cite{Andriluka10cvpr}.
This use case makes it important for these annotations to be of high quality, otherwise they will lead to models fit to their errors. This requires auto-annotation methods to be capable of \emph{self assessment}, i.e. estimating the quality of their own outputs. Self-assessment would allow to return only accurate annotations, automatically discarding the rest.


In this paper we propose a method for transferring knowledge of object appearance from \emph{source} images manually annotated with bounding-boxes to \emph{target} images without them. Source and target images may come from different, but semantically related classes (sec.~\ref{sec:KnowledgeSources}). We model the overlap of an image window with an object using Gaussian Processes (GP)~\cite{RasmussenWilliams05book} regression. GP are able to model highly non-linear dependencies in a non-parametric way. Thanks to probabilistic nature of GP, we are also able to infer the probability distribution of the overlap value for a given window.
This enables taking into account the uncertainty of the prediction.
%
We then tackle the localization problem in a target image by picking a window that has high overlap with an object \emph{with high probability}. The same principle is used for self-assessment.

Doing GP inference directly in a high-dimensional feature space such as HOG~\cite{Dalal05:thomas} is \changed{computationally infeasible on a large scale of ImageNet. Moreover, we would have to estimate thousands of hyper-parameters, risking overfitting to the source.} Instead, we devise a new representation, coined {\em Associative Embedding (AE)}, which embeds image windows in a low dimensional space, where dimensions tend to correspond to aspects of object appearance (e.g. front view, close up, background).
The AE representation is very compact, and embeds windows originally described by HOG~\cite{Dalal05:thomas} or Bag-of-Words~\cite{Zhang07} histograms into \emph{just 3 dimensions}. This enables very efficient learning of GP hyper-parameters and inference. It also facilitates knowledge transfer by allowing the source annotations to spread directly along aspects of appearance.

A large scale experiment on a subset of ImageNet, containing 219 classes and $0.5$ million images, shows that our method outperforms state-of-the-art competitors~\cite{Guillaumin2014ijcv,Guillaumin2012cvpr} and various baselines for object localization. 

The remainder of the paper is organized as follows. The next sec.~\ref{sec:KnowledgeSources} describes the configurations of source and target sets we consider. Sec.~\ref{sec:Overview} formally introduces the setup and gives an overview of our method.
Sec.~\ref{sec:AE} presents Associative Embedding and sec.~\ref{sec:GP} describes how we estimate window overlap with Gaussian Processes. Sec.~\ref{sec:ImplementationDetails} reports some implementation details. Related work is reviewed in sec.~\ref{sec:RelatedWork}, followed by experimental results and conclusion in sec.~\ref{sec:Experiments}.

\parskiny
\section{Knowledge sources}
\label{sec:KnowledgeSources}
\seckiny

\begin{figure}
\begin{center}
\includegraphics[scale=0.2]{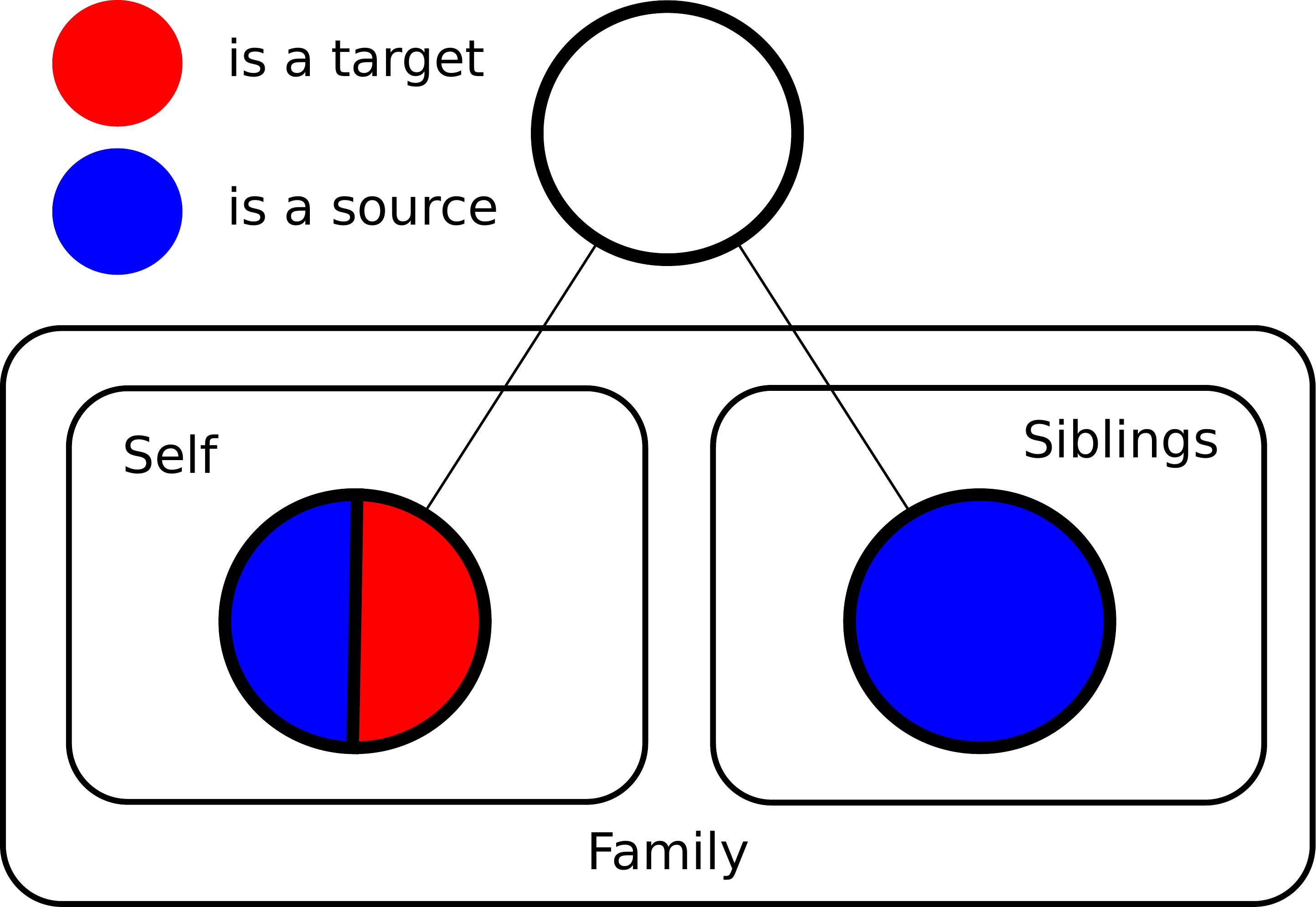}
\figskiny
\end{center}
 \caption{\small{ \em The considered configurations of \textcolor{blue}{\emph{source}} and \textcolor{red}{\emph{target}} sets. Nodes represent classes in ImageNet, connections correspond to "is a" relationships. Plates correspond to different configurations of possible \textcolor{blue}{\emph{sources}}.
 Note that the node of a \textcolor{red}{\emph{target}} class on the left is split, as in certain configurations (self and family) it also contributes some annotated images to the \textcolor{blue}{\emph{source}} set. \figskiny}}
 
\label{fig:Hierarchy}
\end{figure}

The goal of this paper is to localize objects in a target set of images, which do not have manual bounding-box annotations. We tackles this by transferring knowledge from a source set of images that do have them.
The target set is composed of all images of a certain target class without bounding-boxes.
For some classes, ImageNet offers a rather large subset of images with annotations, so both the source and target sets can come from the same class.
In the case when a target class has little or no manual bounding-boxes, we can construct the source set from images of semantically related classes.
Different combinations of source and target sets leads to different learning problems. This paper explores the following setups (fig.~\ref{fig:Hierarchy}):

\begin{enumerate}
\parskiny
\item {\bf Self:} source and target sets of images come from the same class, e.g. from koalas to koalas. This setup is close to a classic object detection problem~\cite{Everingham10}. Here we have the additional knowledge that every target image contains an object of the class and we are given all target images at training time (\emph{transductive learning}).
\parskiny
\item {\bf Siblings:} the source set consists of images from classes semantically related to the target, e.g. from kangaroos to koalas. This is the setup considered in~\cite{Guillaumin2012cvpr}, which can be used to produce annotations even for a target class without any initial manual annotations (\emph{transfer learning}).
\parskiny
\item {\bf Family = self + siblings:} source and target sets consist of images coming from the same mix of semantically related classes, e.g. from kangaroos \emph{and} koalas to kangaroos and koalas. This setup aims at improving performance on related classes when both have some manual annotations by processing them simultaneously (\emph{multitask learning}).

\end{enumerate}
\parskiny
These source/target setups cover a broad range of knowledge transfer scenarios. Below we devise a model that covers all these scenarios in unified manner. We expect an adequate model to have an increasing performance as more supervision is being provided (with \textbf{siblings} having the worst and \textbf{family} having the best performance). 

\parskiny
\section{Overview of our method}
\label{sec:Overview}
\seckiny

\begin{figure*}
\begin{center}
\vspace{-5mm}
\includegraphics[scale=0.18]{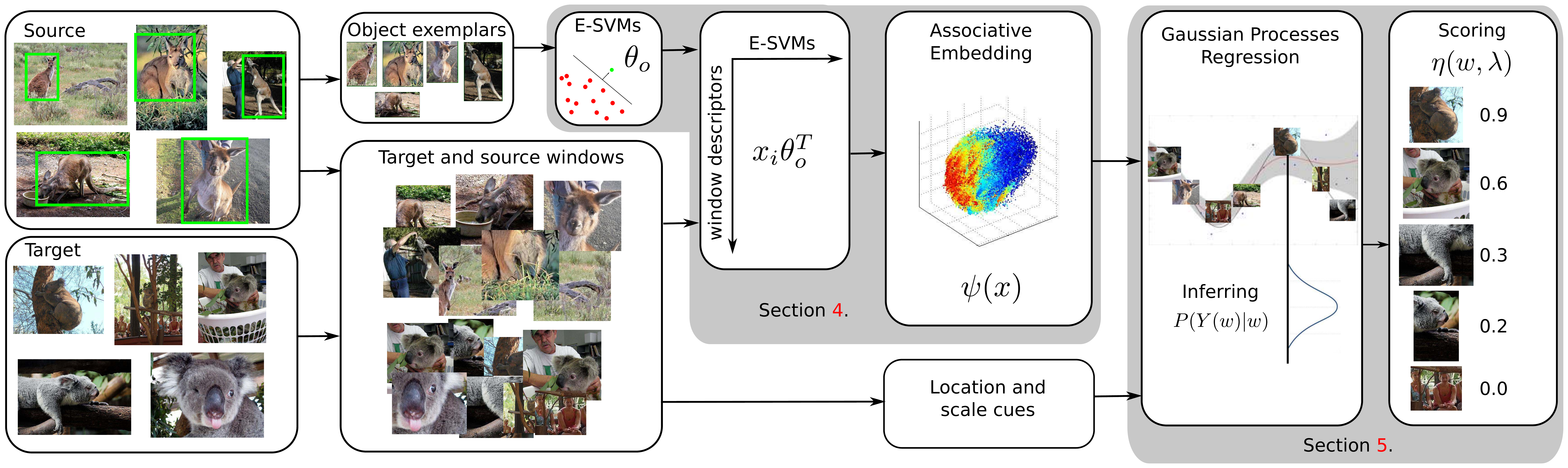}
\vspace{-7mm}
\end{center}
 \caption{\small{  \em Illustration of our pipeline.
 First, E-SVMs are trained for each object exemplar in the source set.
 Next, these E-SVMs are used to represent all \emph{windows} in both source and target sets, which are then embedded into a low dimensional space (coined Associative Embedding, AE).
 After adding location and scale cues to the AE representation of windows, GP regression is used to transfer overlap annotation from source windows to target ones. 
 The final score assigned to a target window integrates the mean predicted overlap and the uncertainty of the prediction. \figskiny}}
\label{fig:Scheme}
\end{figure*}

In this section we define the notation and introduce our method on a high level, providing the details for each part later  (Sec.~\ref{sec:AE},\ref{sec:GP},\ref{sec:LocAndSegm}). Every image $I$, in both source and target sets, is decomposed into a set of windows $\{w_i\}$.
To avoid considering every possible image window, we use the {\em objectness} technique~\cite{alexe12pami} to sample $1000$ windows per image. As shown in~\cite{alexe12pami}, these are enough to cover the vast majority of objects.
Source windows are annotated with the overlap $Y(w)$, defined as their intersection-over-union~\cite{Everingham10} (IoU) with the ground-truth bounding-box of an object. 

The key idea of our method is to use GP regression to transfer overlap annotations from source windows to target ones. GP infers a probability distribution over the overlap of a target window with the object: $P(Y(w^t)=y|w^t)$~\footnote{We slightly abuse notation here, dropping the conditioning on source data: $P(Y(w^t)=y | w^t,\{Y(w_s) = y_s\}_{s\in \mathtt{source}})$}. 
Having a full distribution enables to measure and control uncertainty. Consider the score function\footnote{Corrected thanks to O. Russakovsky}:
\vspace{-2mm} 
\begin{equation}
\eta(w,\lambda) \equiv \max_{\xi \in  \left[0,1\right]  }{\xi} \; \; \; \; \mathtt{         s.t.} \; P(Y(w) \geq \xi|w) \geq \lambda .
\vspace{-3mm}  \label{eq_score}
\end{equation}
This score is the largest overlap $Y(w)$ that a window $w$ is predicted to have with at least $\lambda$ probability.
The higher the $\lambda$, the more certain we need to be before assigning a window a high score. 
 
Performing GP inference in the high-dimensional space typical for common visual descriptors (e.g. HOG, bag-of-words) \changed{is computationally infeasible at the large scale of ImageNet. Moreover, it would require estimating thousands of hyper-parameters, risking overfitting.} To reduce the dimensionality of the feature space without losing descriptive power, we introduce Associative Embedding (AE) $\psi$, a low-dimensional representation $\dim(\psi(x)) << \dim(x)$ of windows (sec.~\ref{sec:AE}).
Let $x_o$ be the representation of an object \emph{exemplar} (ground-truth bounding-box) from the source set in some feature space (e.g. HOG).
Let $x_i$ the representation of a window $w_i$ in the same feature space. First we describe $w_i$ in terms of its similarity to the exemplars $\{x_o\}_{o=1}^{N_o}$ in the source set.
We quantify how similar the window $x_i$ is to the exemplar $x_o$ by the output of an Exemplar-SVM (E-SVM)~\cite{MalisiewiczICCV11} with parameters $\theta_o$ trained on $x_o$.
Now every window is described by a vector of E-SVM outputs $[x_i \theta_1^T, ..., x_i \theta_{N_o}^T] $, one for each exemplar in the source set. Intuitively, this intermediate representation describes what a window \emph{looks like}, rather than \emph{how it looks}. The advantage of using E-SVMs over a standard similarity measure is its ability to suppress the influence of the background present in the window and to focus on exemplar-specific features~\cite{MalisiewiczICCV11}.

Next, we embed all windows and exemplars into a low dimensional space. We optimize the embedding to minimize the difference between scalar products in the original and embedded spaces. We call the resulting space {\em Associative Embedding} (AE). The dimensions in EA space tend to correspond to aspects of the appearance of classes (e.g. close up, side view, partially occluded). The AE space enables the GP to transfer knowledge directly along the aspects which greatly facilitates the process. For HOG and Bag-of-Words descriptors, an AE space with \emph{just 3 dimensions} is sufficient to support this process.

Fig.~\ref{fig:Scheme} depicts our method procedural flow. Having embedded source and target windows into AE space, we use GP to infer probability distributions of target windows overlap $P(Y(w^t)=y|w^t)$ and to compute the score $\eta(w^t,\lambda)$ (eq.~(\ref{eq_score})). We employ this score to tackle localization task in sec.~\ref{sec:LocAndSegm}.
For localization we simply return the highest scored window in each target image. We also reuse the score for self-assessment.

\parskiny
\section{Associative embedding}
\seckiny
\label{sec:AE}

\begin{figure}
\begin{center}
\includegraphics[scale=0.15]{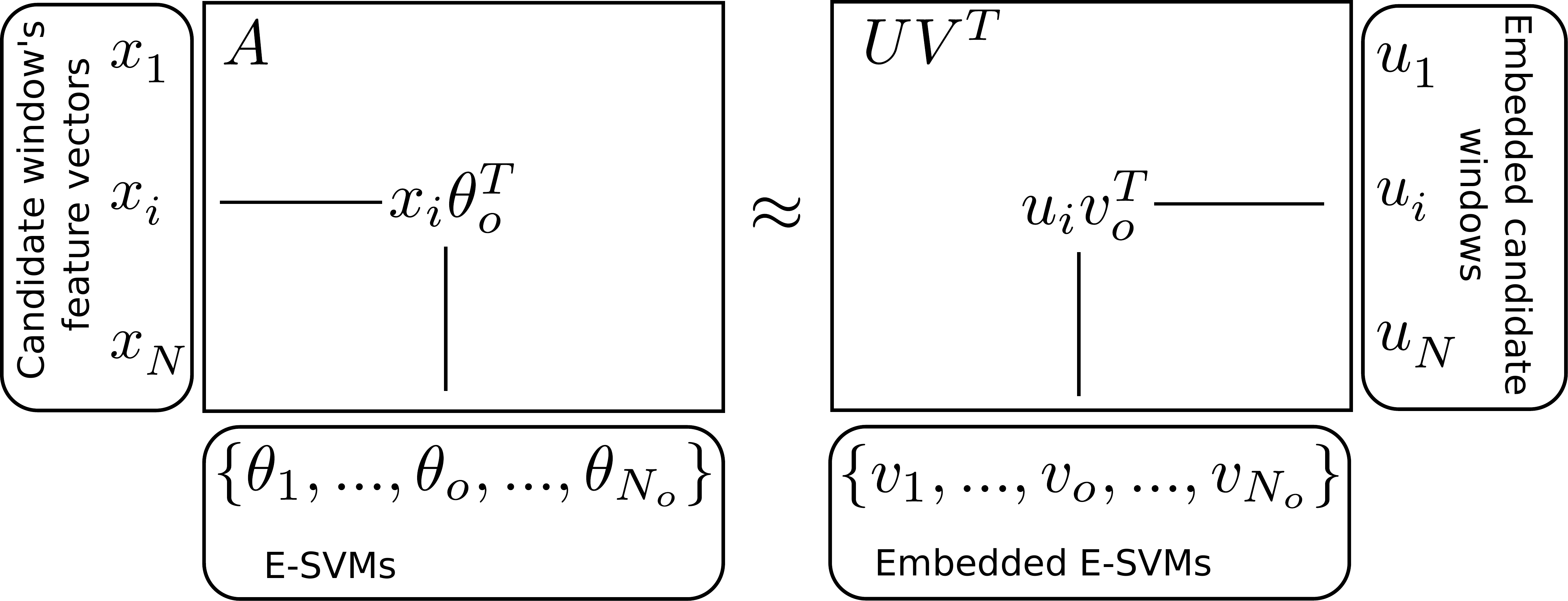}
\end{center}
\figskiny
 \caption{\small{ \it Illustration of Associative Embedding (AE). A matrix of E-SVM outputs $A$ (left) on windows from both source and target sets is factorised into $UV^T$ (right). \figskiny}}
\label{fig:AE_illustr}
\end{figure}

This section details the proposed AE representation (fig.~\ref{fig:AE_illustr}). The method proceeds by first training E-SVMs for object exemplars in the source set, then representing  all windows from both source and target sets as the output of E-SVMs applied to their feature descriptors, and finally embedding these representations in a lower dimensional space. 

\parshrinky
\paragraph{Training E-SVMs.}
For each object exemplar $x_o$ in a given feature space, we train an E-SVM~\cite{MalisiewiczICCV11} hyperplane $\theta_o$ to minimize the following loss
\begin{equation}
\theta_o = \argmin_{\theta}{ ||\theta||^2 + C_1 h(x_o \theta^T) - C_2 \sum_{x_i \in {\cal N}} {h ( x_i\theta^T ), }}
\end{equation}
where $h$ is the hinge loss $h(x) = \max{(0,1-x)}$ and ${\cal N}$ is a set of negative windows from the source pool. It contains a few thousand windows with overlap smaller than 0.5 with the object's bounding-box.
The weights $C_1$ and $C_2$ regulate the relative importance of the terms.

\parshrinky
\paragraph{Encoding windows as E-SVM outputs.}
We encode every window $w_i$ in both the source and target sets as a vector $a_i$ where each element $a_{io}=x_i \theta_o^T$ is the response of an E-SVM $\theta_o$. 
This representation is much richer than the original feature space, as each entry in $a_i$ encodes how much this window looks like one of the exemplars.

\begin{figure}
\begin{center}
\includegraphics[scale=0.2]{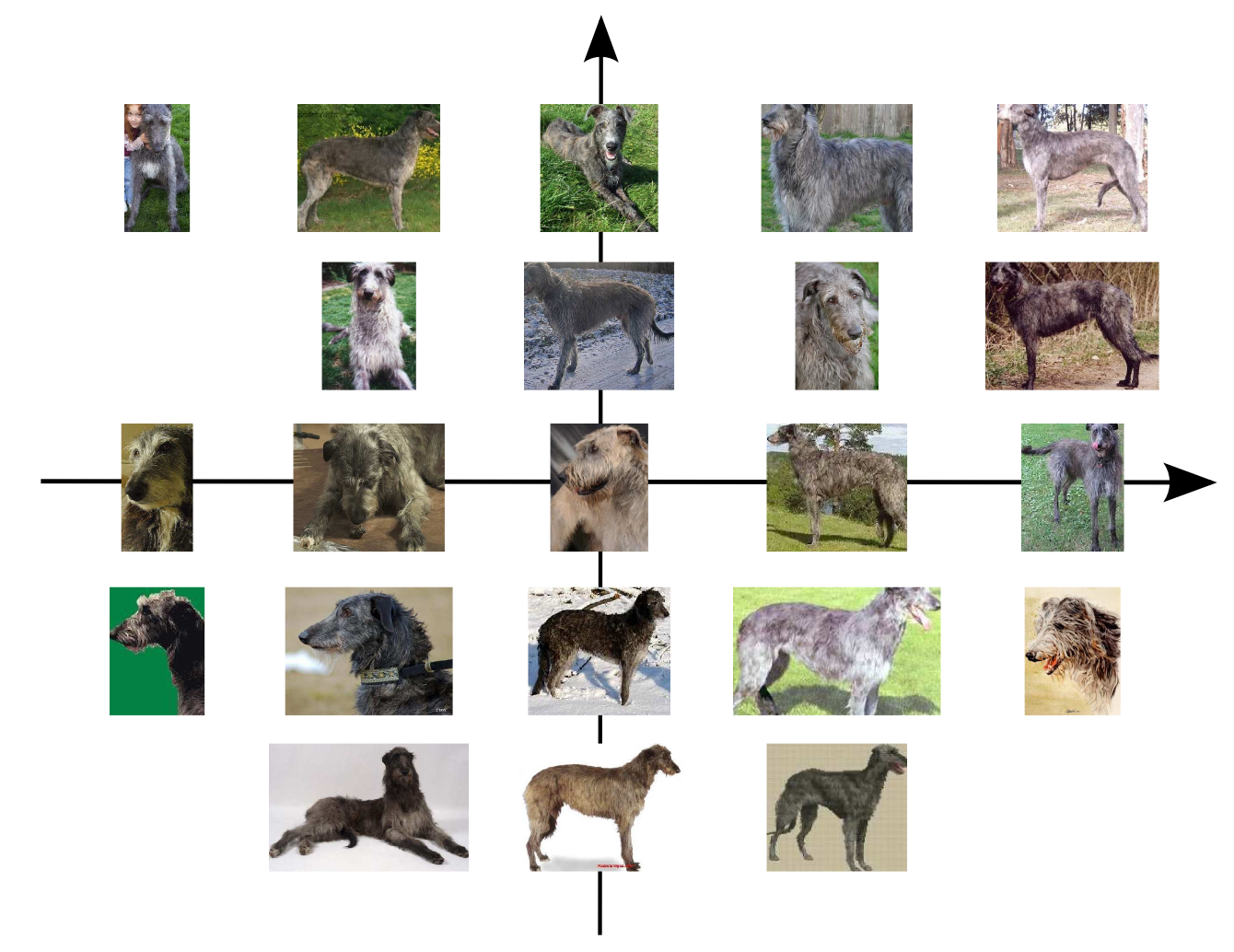}
\end{center}
\figskiny
 \caption{\small{ \it A visualization of AE for the class `Scottish Deerhound'. 
 We show some source windows with high overlap with the ground-truth bounding-box. The position corresponds to the 2nd and 3rd coordinates in their AE representation. Note how windows with similar appearance cluster together: the top right corner corresponds to facing-left side views, the center to face close ups, and the bottom to facing-right side views. This shows how AE manages to recover the underlying appearance variation dimensions.\figskiny}}  
\label{fig:Dogs_embedded}
\end{figure}

\parshrinky
\paragraph{Embedding.}
Let $A$ be a matrix with a row $a_i$ for each window in both the source and target sets.
Let $u_i$ and $v_o$ be the embedded representations of $x_i$ and $\theta_o$, which we are trying to produce.
Let matrices $U$ and $V$ have $u_i$ and $v_o$ as rows, respectively.
We seek the embedding that minimizes the reconstruction error:
\begin{equation}
\min_{U,V} ||U V^T - A||_F^2
\label{eq:embedding}
\end{equation}
where $||\cdot ||_F$ is the Frobenius norm.
This can be done by a truncated SVD decomposition~\cite{Deerwester90}: $A=U \Sigma \hat{V}^T$. We encapsulate singular values into the E-SVM representation: $V=\Sigma \hat{V}^T$.  
Elements of a common visual descriptor (e.g. HOG) correspond to local statistics of an image window. However, each element of $a_i$ correspond to the output of a battery of E-SVMs on the same image window. \changed{These are individually much more informative, yet their collection is redundant, enabling compression to a few dimensions. 
For instance, the representation $a_i$ of a background window will contain similar negative values across all the E-SVMs. Moreover, E-SVMs that are learnt from similar exemplars coming from the same aspect of object's appearance (e.g., gorilla's face, close up) will produce correlated outputs uniformly for all windows.
Altogether, this dramatically collapses $a_i$ variability, which} allows SVD to achieve low reconstruction error with just a few dimensions, which tend to correspond to object/background discrimination and the aspects of object appearance (fig.~\ref{fig:Dogs_embedded}).


Solving the optimization problem for all windows in the source and target sets at once is computationally very expensive.
Therefore, in a first step we use only a sub-sample of the windows, but all exemplars, to minimize (\ref{eq:embedding}). This results in an embedded representation of all exemplars $V$.
In a second step we keep $V$ fixed and embed any remaining window $x_i$ with
\begin{equation}
\psi(x_i) = \argmin_u{ \sum_{o \in \mathrm{source}} { || x_i \theta_o^T - uv_o^T ||^2 }}
\end{equation}

We solve the above optimization using least-squares. Fig.~\ref{subfig:Koala_cloud} depicts the embedding of Koala class SURF Bag-of-Words window descriptors in AE space.

\parskiny
\section{Estimating window overlap with Gaussian Processes}
\label{sec:GP}
\seckiny
This section describes how to infer a probability distribution $P(Y(w^t)|w^t)$ of the overlap of a target window $w^t$ with an object and calculate the final score (eq.~\ref{eq_score}). 
We first construct an extended representation of a window using AE, its objectness score~\cite{Alexe10cvpr} and its position and scale in the image.
Then we define a GP over that representation and estimate its hyper-parameters. We also show how to speed up inference at prediction time in sec.~\ref{sec:LargeScaleGP}.

\parskiny
\subsection{GP construction}
Let $x^{\mathrm{SURF}}$ and $x^{\mathrm{HOG}}$ be the SURF bag-of-words and HOG appearance descriptors of a window $w$, and $\psi(x^{\mathrm{SURF}}), \psi(x^{\mathrm{HOG}})$ their AE. The final representation $\phi(w)$ of $w$ is
\begin{equation}
\small
\phi(w) = [c(w),\mathrm{Obj}(w), \psi(x^{\mathrm{SURF}}), \psi(x^{\mathrm{HOG}})],
\end{equation}
where $c(w)$  is a position and scale descriptor of a window. Following~\cite{Guillaumin2012cvpr} we define it as $c(w) = [c_1,c_2, \log(WH),\log(W/H)]$. Here $c_1$ and $c_2$ are the coordinates of the center of the window, $W$ and $H$ are width and height of the window (all normalized by the image size). $\mathrm{Obj}(w)$ is the objectness score of $w$, which estimates how likely it is to contain an object rather than background~\cite{alexe12pami}.

We propose to model the overlap $Y(w)$ of a window $w$ with a true object bounding-box as a Gaussian Process (GP)
\begin{equation}
Y(w) \sim {\cal GP}\left( m(\phi(w)), k(\phi(w),\phi(w')) \right)
\label{eq:GP_window_score}
\end{equation}
where $m(\phi(w))$ is a mean function and $k(\phi(w),\phi(w'))$ is a covariance function.  Here the mean function is zero $m(\phi(w))=0$ and the covariance (kernel) $k( \phi(w), \phi(w'))$  is the squared exponential
\begin{equation}
\exp\left( -\frac{1}{2}
(\phi(w) - \phi(w'))^T \text{diag}(\gamma)^{-2}
(\phi(w) - \phi(w')) \right)
\end{equation}
where $\gamma$ is a vector of hyper-parameters regulating the influence of each element of $\phi(w)$ on the
output of the kernel.

Let $\{w_{s},y_{s}\}^{N_s}_s$ be a set of windows from the source set and their respective ground-truth overlaps $y_s = Y (w_s)$ (\emph{inducing points}). Let $K$ be a kernel matrix $K_{s,s'}=k( \phi(w_s), \phi(w_{s'}))$.
Consider now a target window $w^t$, for which $Y(w^t)$ is unknown.  Let $\bm{k} = \left[ k(w^t , w_1), k(w^t , w_2), ..., k(w^t , w_{N_s}) \right]$. Then a predictive distribution for $w^t$ is
\begin{equation}
P(Y(w^t)|w^t) = {\cal N} \left( \mu(w^t),\sigma^2 (w^t) \right)
\end{equation}
where
\[
\mu(w^t) = \bm{k}^T K^{-1} \bm{y}_{1:N}
\]
\begin{equation}
\sigma^2 (w^t) = k(w^t, w^t) - \bm{k}^T K^{-1} \bm{k}
\end{equation}

In practice, inference involves the inversion of kernel matrix $K$ and a series of scalar product. Matrix inversion for a large set of inducing points poses a computational problem, with which we deal with in sec.~\ref{sec:LargeScaleGP}.

\begin{figure*}
\begin{center}
    \subfigure[\footnotesize{\vspace{-7mm} }] {\label{subfig:Koala_cloud}\includegraphics[scale=0.32]{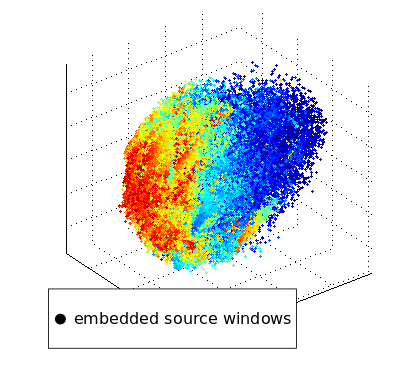}}            \hspace*{22pt}
    \subfigure[\footnotesize{\vspace{-7mm}}] {\label{subfig:Koala_cloud_with_testImage}\includegraphics[scale=0.25] {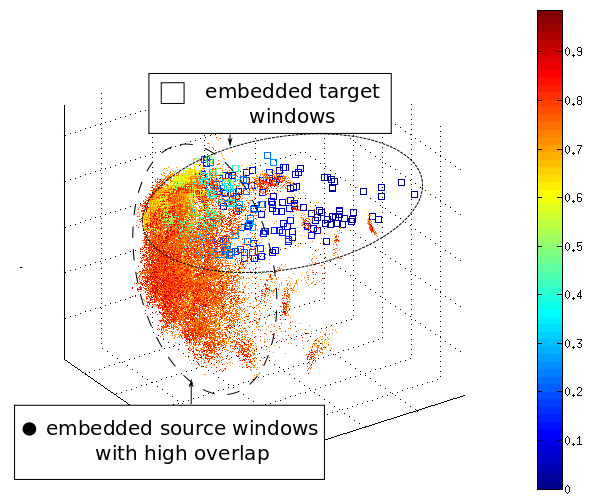}} \hspace*{22pt}       
    \subfigure[\footnotesize{\vspace{-7mm}}] {\label{subfig:Koala_image_with_boxes}\includegraphics[scale=0.3] {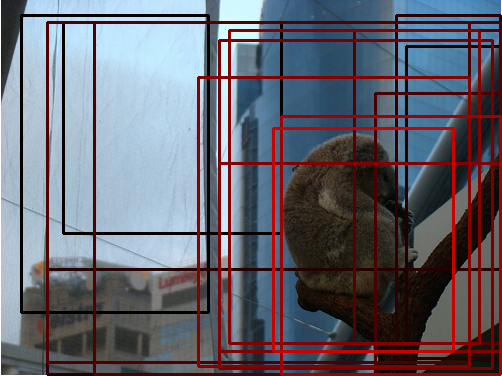}}
\end{center}
\figskiny
\caption{\small{\it 
Illustration of AE of windows of the Koala class, using the SURF Bag-of-Words descriptor. 
(a) the embedded source windows. Points correspond to windows and their colour to the overlap with the ground-truth objects in their respective images.
Note how source windows form a comet shape with background windows in the center and the tail of the comet.
(b) the embedded target windows of an image (squares) together with source windows (dots) with high overlap with the ground-truth.
(c) a target image with a few randomly sampled windows. Colour intensity corresponds to the overal score $\eta(w,0.5)$ output by our method. \figskiny  }}

\end{figure*}

\parshrinky
\paragraph{Overlap estimation.}
We can now compute the score $\eta(w,\lambda)$ from eq.~(\ref{eq_score}).
\begin{equation}
\eta(w,\lambda) = \mu(w) + \beta(\lambda)\sigma (w)
\label{eq:GP_score}
\end{equation}
where $\beta$ is a constant which depends on $\lambda$ and can be analytically computed by integrating the Gaussian density. The score $\eta(w,\lambda)$ equals to the maximum overlap that window $w$ may have with at least $\lambda$ probability. 

\parshrinky
\paragraph{Estimating hyper-parameters.}
We estimate the hyper-parameters $\gamma$ by minimizing the regularized negative log-likelihood of the overlaps of the windows in the source set
\begin{equation}
-\log P \left( {\cal D}(w) | \phi(w), \gamma \right) + \alpha ||\gamma||^2,
\end{equation}
where $\alpha$ is the regularization strength. While it is not common practice to put a regularizer on $\gamma$, we experimentally found that it significantly improves the result.

\subsection{Fast inference for large-scale datasets}
\label{sec:LargeScaleGP}
\vspace{-2mm}
The source set can contain millions of windows, which poses a computational problem for standard GP inference methods~\cite{RasmussenWilliams05book}. Instead of exact inference we use an approximation~\footnote{infFITC from the toolbox~\cite{rasmussen2010ijcv} } during both training and test. For training hyper-parameters we also sub-sample the windows from the source set. Since the dimensionality of $\gamma$ is very low ($11$ in our case), it can be reliably estimated from modest amounts of data.

To speed up prediction on windows in a target image, as inducing points of GP we only use windows from source images that have a similar global appearance. The idea is that globally similar images are more likely to contain objects and backgrounds related to those in the target image, and therefore are the most relevant source for transfer. This is related to ideas proposed before for scene parsing~\cite{liu2009nsp,tigheECCV10}.
However, while those works used a simple monolithic global image descriptor (bag-of-words or GIST) for this task, we directly use the set of window descriptors $\phi(w)$ to construct the global similarity measure. In the feature space spanned by $\phi$, every image has an empirical distribution, i.e. a cloud of points corresponding to the windows it contains (fig.~\ref{subfig:Koala_cloud_with_testImage}). We use per-coordinate kernel density estimation to represent the cloud of an image. The global image similarity between a source and a target image is then defined as the average per-coordinate KL-divergence between their distributions. 

All the modifications above reduces full inference for all windows in a test image to approximately 4 seconds on a standard desktop computer. Hence our method is suitable for large scale datasets like ImageNet.

\parskiny
\section{Object localization}
\seckiny
\label{sec:LocAndSegm}

The technique presented above produces a score $\eta(w,\lambda)$ for each target window in a target image $w \in I^t$. This section explains how to use this score for object localization. 
We simply select the window with the highest score $w^*=\argmax_{w \in I^t}{\eta(w,\lambda)}$, out of all windows in the target image. This window is the final output of the system, which returns one window for each target image.

The score can be also be used for self-assessment. For example, we can retrieve all bounding-boxes that have overlap higher than $60\%$ with $0.8$ probability by taking only the windows such that $\eta(w,0.8)>60$.




\parskiny
\section{Implementation details}
\seckiny
\label{sec:ImplementationDetails}

\paragraph{Associative Embedding.}
We use AE in SURF~\cite{Bay08} bag-of-words~\cite{Zhang07} and HOG feature spaces. We build quantized SURF histograms with a codebook of $2000$ visual words. For computational efficiency, we approximate the $\chi^2$ kernel with the expansion technique of~\cite{vedaldi10cvpr} and train linear E-SVMs.
HOG descriptors~\cite{Dalal05:thomas} are computed over a $8\times 8$ grid and the associated E-SVMs are linear, as in~\cite{MalisiewiczICCV11}.
The soft margin parameters $C_1$ and $C_2$ are set to $C_1=1$ and $C_2=0.001$. We create AE of a dimensionality $3$ (for each of HOG and SURF bag-of-words).
\parshrinky
\paragraph{Gaussian Processes.}
To learn GP hyper-parameters (sec.~\ref{sec:GP}) we sub-sample $15000$ windows from the source set. Since the dimensionality of GP hyper-parameters is low ($\dim( \gamma ) = 11$ in our experiments), this is sufficient.
We set the regularizer to $\alpha=100$ in eq. (\ref{eq:GP_score}).
For the scoring function $\eta(w,\lambda)$ we set $\lambda=0.8$ for object localization, and $\lambda=0.5$ for self-assessment.

For prediction on a target image, we retrieve the $300$ most similar source images as described in sec.~\ref{sec:LargeScaleGP}, and then infer the distribution of windows overlap using them. 

\parskiny
\section{Related work}
\seckiny
\label{sec:RelatedWork}

\paragraph{Populating ImageNet.} The two most related works are~\cite{Guillaumin2014ijcv,Guillaumin2012cvpr}. Like us, they address the problem of populating ImageNet with automatic annotations (bounding-boxes~\cite{Guillaumin2012cvpr} and segmentations~\cite{Guillaumin2014ijcv}). 
Guillaumin and Ferrari (GF)~\cite{Guillaumin2012cvpr} train a series of monolithic window classifiers, one for each source class, using different cues (HOG, colour, location, etc.). They are combined into a final window score by a discriminatively trained weighting function and applied to windows in the target set.
As we show in sec.~\ref{sec:Experiments}, GF is not suited for self assessment. Moreover, our method produces better localizations overall.

The work~\cite{Guillaumin2014ijcv} populates ImageNet with segmentations. It propagates ground-truth segmentations from PASCAL VOC~\cite{Everingham10} onto ImageNet. They use a nearest neighbour technique~\cite{kuettel12cvpr} to transfer segmentations from a given source set to a target image. 
\changed{We compare to this method experimentally (sec.~\ref{sec:Experiments}), by  putting a bounding-box over their segmentations.}

\parshrinky
\paragraph{Transfer learning} is used in computer vision to facilitate learning a new target class with the help of labelled examples from related source classes. Transfer is typically done through regularization of model parameters~\cite{fei2007CVIU,tommasi2010cvpr}, an intermediate attribute layer~\cite{lampert:cvpr09} (e.g. yellow, furry), or by sharing parts~\cite{ott11cvpr}.
In GP~\cite{bonilla2008nips} transfer learning is usually based on sharing hyper-parameters between tasks. In this work we not only share hyper-parameters, but the \emph{inducing points} as well. Also, our GP kernel is defined over an augmented AE space $\phi(x)$, which is constructed specifically for a particular combination of source and target classes. In principle, one could view AE as kernel learning method for GP, which exploits the specifics of visual data.

\parshrinky
\paragraph{Exemplar SVMs} were first proposed by~\cite{MalisiewiczICCV11} and are rapidly gaining popularity as a better way to measure visual similarity~\cite{endres2013cvpr,aghazadeh2012eccv,Dong13cvpr}.
Their main advantage is the ability to select features within a window that are relevant to the object, down-weighting background clutter. Some authors propose to use~\cite{aghazadeh2012eccv,Dong13cvpr} E-SVMs as a similarity measure for discovering different aspects of object appearance. They  explicitly group training objects into clusters according to their aspects of appearance. They produce a set of clean aspect-specific object detectors, whose responses on a test image are merged together for the final result. 
Instead we embed image windows into a low dimensional space, where aspects of appearance are expressed smoothly in the space's dimensions. 


 \parskiny
\section{Experiments and conclusion}
\label{sec:Experiments}

\begin{figure}
\begin{center}
\includegraphics[scale=0.9]{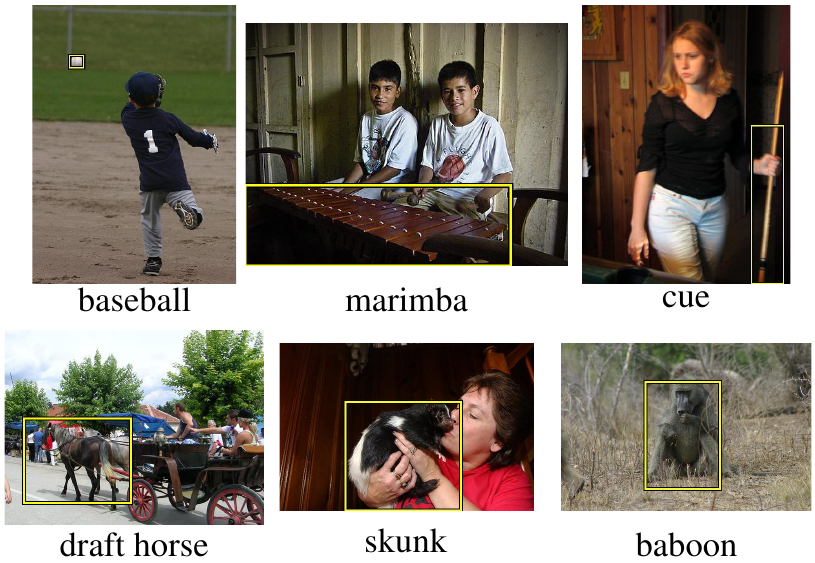}
\end{center}
\figskiny 
 \caption{\small{ \it Localization results for our AE-GP+ method. \figskiny \figskiny \figskiny}} \figskiny 
\label{fig:Detection_vsMTH}
\end{figure}

We perform experiments on the same subset of ImageNet as~\cite{Guillaumin2014ijcv,Guillaumin2012cvpr}, which allows direct comparison to their results.
The subset consists of 219 classes (e.g. phalanger, beagle, etc.) spanning over $0.5$ million images. Classes are selected such that each has less than $10.000$ images and has siblings with some manual bounding-box annotations~\cite{Guillaumin2012cvpr}.
For a given target class we consider the following source sets:
i) \textbf{self} --- the class itself;
ii) \textbf{siblings} of the class (as in~\cite{Guillaumin2012cvpr});
iii) \textbf{family} --- the class itself plus its siblings (see sec.~\ref{sec:KnowledgeSources} and fig.~\ref{fig:Hierarchy}).

\parshrinky
\paragraph{Ground-truth.} We use 92K images with ground-truth bounding-boxes in total. We split them in two disjoint sets of 60K and 32K. We use the first set exclusively as source and the second one exclusively as target. This allows us to compare transfer from different source types (siblings, self, family) on exactly the same target images.
The first batch of 60K images is the same as used in~\cite{Guillaumin2012cvpr}. This ensures proper comparison to~\cite{Guillaumin2012cvpr}, as when using siblings as source, our method transfers knowledge from exactly the same images as~\cite{Guillaumin2012cvpr}.
%


\begin{figure}
\begin{center}
\includegraphics[scale=0.4]{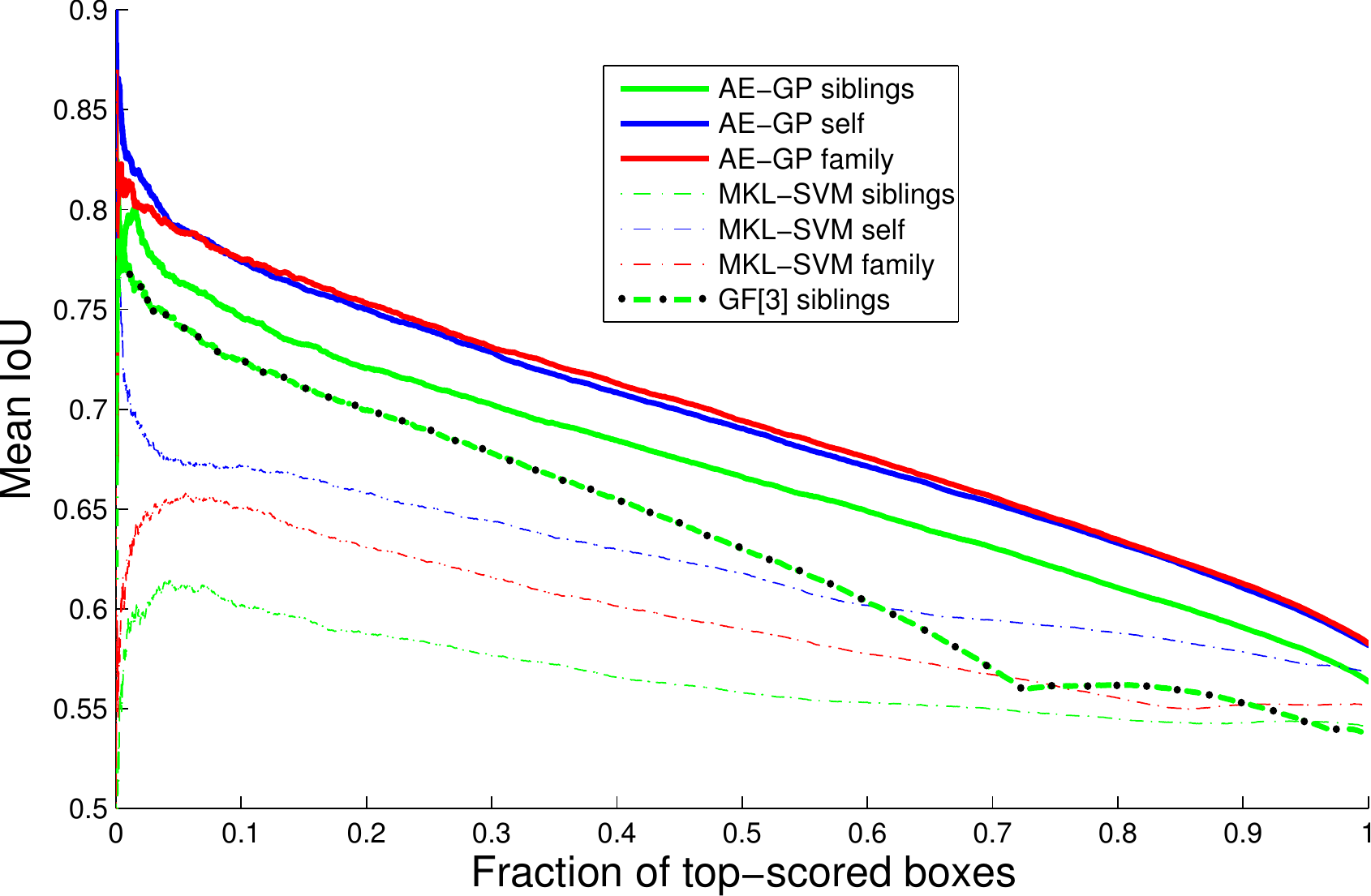}
\end{center}
\figskiny
 \caption{\small{\em Self-assessment performance curves for our method (AE-GP), GF~\cite{Guillaumin2012cvpr}, and MKL-SVM. \figskiny  \vspace{-2mm} }}
\label{fig:All_curves}
\end{figure}

\parshrinky
\paragraph{Baselines and \cite{Guillaumin2012cvpr}.}
For localization, we compare against the following.
%
{\textbf{MidWindow:} A window in the center of the image, occupying $50\%$ of its area.}
%
\textbf{TopObj:} The window with the highest objectness score~\cite{alexe12pami}.
%
\textbf{MKL-SVM:}
This represents a standard, discriminative approach to object localization, similar to~\cite{Vedaldi09}. On the source set we train a SVM on HOG (linear) and a SVM with SURF bag-of-words (linear with $\chi^2$ expansion~\cite{vedaldi10cvpr}) using $90\%$ of the data. To combine them, we train a linear SVM over their outputs using the $10\%$ holdout data.
We improve the baseline by adding the objectness score, location and scale cues (as in sec.~\ref{sec:GP}) of a window as features for the second-level SVM. 
\textbf{GF:} We compare to~\cite{Guillaumin2012cvpr}, when using siblings as the source. We use the output of~\cite{Guillaumin2012cvpr} as provided by the authors on their website. \changed{Notice how MKL-SVM baseline and the competitor GF~\cite{Guillaumin2012cvpr} are defined on the same object proposals~\cite{alexe12pami} and features as our AE-GP. This ensures that any improvement comes from better modelling.}

\parshrinky
\paragraph{Metrics.}
To measure the quality of localizations we use the {\em intersection-over-union} criterion (IoU) as defined in the PASCAL VOC~\cite{Everingham10}.
We also measure {\em detection rate}: the percentage of images where the output has IoU $>0.5$.

To measure the quality of self-assessment we evaluate how well the score (\ref{eq_score}) ranks the output bounding-boxes. We sort the outputs by their scores and measure the mean IoU of the top $p\%$ outputs.
To compare to~\cite{Guillaumin2012cvpr} we use their scores released along with their output bounding-boxes. For MKL-SVM we use the score output by the second-level SVM. 

\parshrinky
\paragraph{Comparison to baselines and \cite{Guillaumin2012cvpr}.}


Table~\ref{table:results} summarizes localization results, averaged over all 32K target images for which we have evaluation ground-truth.
The results of our method steadily improve as the source set changes from siblings to self to family, unlike the MKL-SVM baseline, whose performance decreases from self to family. This behaviour shows that our method is applicable to a wide range of knowledge transfer scenarios. Using only siblings as source, we outperform GF by the same margin as GF outperforms the trivial baselines TopObj and MidWindow in terms of mean IoU. 
Overall, our method delivers the best results for all kinds of source sets and metrics, compared to both competitors and baselines. 
While GF~\cite{Guillaumin2012cvpr} sometimes produces a bounding-box occupying most of an image, our localizations are typically more specific to the object.


Fig.~\ref{fig:All_curves} presents self-assessment curves.
Our method nicely trades off the amount of returned localizations for their quality, as demonstrated by the visible slope of the solid curves. For all sources, our method outperforms MKL-SVM and GF over the entire range of the curve.
The advantage over MKL-SVM is greater especially in the left part of the curves, where self-assessment plays a bigger role. 
Note how both MKL-SVM and GF have cusps in their curves. This means that many high quality localizations get a low score (GF, right half of the curve) or some high scoring localizations are poor (MKL-SVM, left half).
Interestingly, our MKL-SVM baseline performs similarly to GF when evaluated on all images (tab.~\ref{table:results}), but GF is better at self-assessment (fig.~\ref{fig:All_curves}). 

\begin{table}
\begin{center}
{\scriptsize
\begin{tabular}{c|c|c|c|c}
Method & Source & IoU\% & IoU at 50\% & Detection\%\tabularnewline
\hline 
\hline 
AE+GP & Siblings & \textbf{56.4} & \textbf{66.6} & \textbf{63.5}\tabularnewline
\hline 
GF~\cite{Guillaumin2012cvpr} & Siblings & 53.7 & 63 & 58.5\tabularnewline
\hline 
MKL-SVM & Siblings & 54.1 & 55.8 & 59.7\tabularnewline
\hline 
\hline 
AE+GP & Self & \textbf{58.2} & \textbf{69} & \textbf{66.5}\tabularnewline
\hline 
AE+GP+ & Self & \textbf{59.9} & \textbf{71} & \textbf{68.3}\tabularnewline
\hline 
MKL-SVM & Self & 56.7 & 61.8 & 63.8\tabularnewline
\hline 
\hline 
AE+GP & Family & \textbf{58.3} & \textbf{69.4} & \textbf{66.7}\tabularnewline
\hline 
MKL-SVM & Family & 55.1 & 58.9 & 60.8\tabularnewline
\hline 
\hline 
TobObj & - & 50.1 & 55.3 & 52.7\tabularnewline
\hline 
{MidWindow} & - & 49.3 & - & 60.7\tabularnewline
\hline 
{KGF~\cite{Guillaumin2014ijcv}} & - & 59.9 & - & 66.9\tabularnewline
\end{tabular}
}
\end{center}
\vspace{-3mm} 
 \caption{\small{ \it Localization results: mean IoU, mean IoU of the top $50\%$ ranked outputs, and detection rate. \vspace{-6mm} }\label{table:results}}
\end{table}

\begin{figure}
\begin{center}
\includegraphics[scale=0.46]{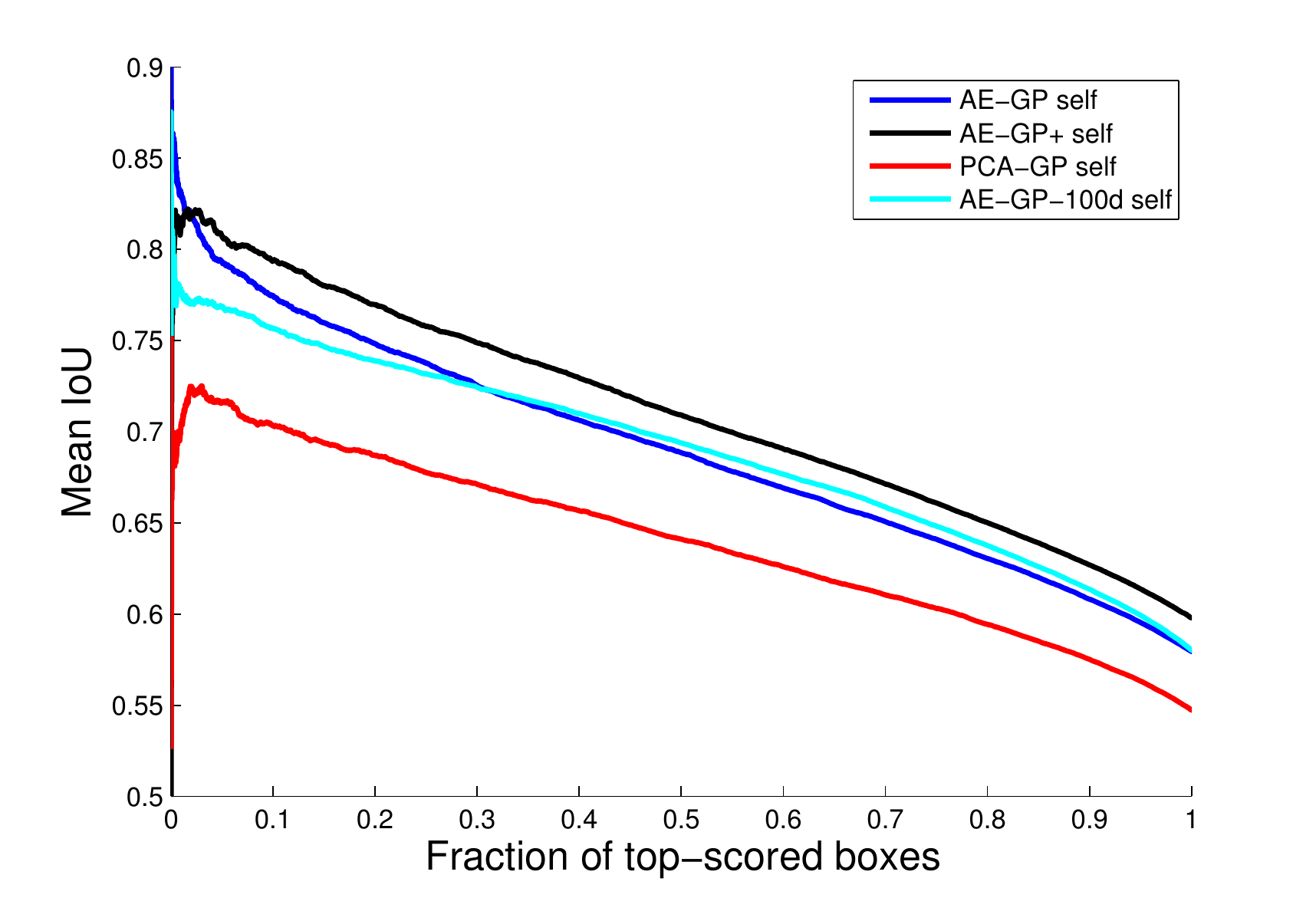}
\end{center}
\figskiny
 \caption{\small{\em Self-assessment performance curves for AE-GP versions. \figskiny \figskiny }}
\label{fig:Extra_curves}
\end{figure}

\parshrinky
\paragraph{Analysis of AE.}
\changed{We validate here the importance of AE and whether it can reduce dimensionality without loss in representation power (fig.~\ref{fig:Extra_curves}). }

\changed{In \textbf{PCA-GP} we substitute AE with a standard PCA in each original feature space (HOG, SURF bag-of-words), keeping the rest of the method exactly the same. PCA-GP performs significantly worse than AE-GP, highlighting the importance and power of AE. }

\changed{\textbf{AE-GP-d100} increases the dimensionality of AE to $100$ for each feature space. This drastic increase of dimensionality makes little difference in the results. This shows that AE does not lose much representation power even when reducing the dimensionality to just 3.}

\parshrinky
\paragraph{Better features and proposals.}
\changed{ The experiments above demonstrate that our AE-GP outperforms baselines and \cite{Guillaumin2012cvpr}, given the same features and object proposals. To push performance even further we introduced here a version of our method, coined \textbf{AE-GP+}, that uses state-of-the-art features~\cite{uijlings2013ijcv} and object proposals~\cite{Manen2013iccv}. To accommodate very high dimensional features~\cite{uijlings2013ijcv} we slightly increase the dimensionality of AE to $10$. We use the initial features (sec.~\ref{sec:ImplementationDetails}) as well, improving SURF bag-of-words by adding spatial binning. Results in fig.~\ref{fig:Extra_curves} and tab.~\ref{table:results} show that AE-GP+ delivers the best results, improving over AE-GP by $1.8\%$ in detection rate. Fig.~\ref{fig:Detection_vsMTH} demonstrates localizations by AE-GP+.}

\parshrinky
\paragraph{Comparison to \cite{Guillaumin2014ijcv}.}
\changed{We compare here to the state-of-the-art segmentation method KGF~\cite{Guillaumin2014ijcv} by putting a bounding-box around their output segmentation. AE-GP+ moderately outperforms KGF~\cite{Guillaumin2014ijcv} by $1.4\%$ in detection rate (tab.~\ref{table:results}). Most importantly, as KGF~\cite{Guillaumin2014ijcv} assigns no score to its output, self-assessment is impossible. 
The user can't \emph{automatically} retrieve high quality localizations from KGF~\cite{Guillaumin2014ijcv}. In contrast, AE-GP+ has the ability to select the localizations that have high IoU.
Also, unlike the scoring schemes in GF and MKL-SVM, ours allows the user to retrieve localizations that are \emph{predicted} to have an overlap higher than, say, $60\%$ with a $0.5$ probability. Using AE-GP+ with self as source, this fully automatically returns $51\%$ of all localizations with
a mean IoU of $71\%$ (which is very accurate, c.f. the PASCAL detection criterion is IoU$>0.5$).
This means about 251K images in the dataset we processed! The results are released online~\footnote{ {\footnotesize{http://groups.inf.ed.ac.uk/calvin/proj-imagenet/page/}} }.}

\parshrinky
\paragraph{Conclusion.}

{Knowledge transfer in ImageNet is motivated by the fact that semantically related objects \emph{look similar} (e.g. police car and taxi). Hence, in a latent space of appearance variation image windows that contain objects of related classes are close to each other, while background windows are far away from them.
Our work formalizes this intuition with the AE+GP method.} AE recovers the latent space of appearance variation and embeds windows into it. Next, we construct a GP over the AE space to transfer localization annotations from the source to target image set.
Thanks to probabilistic nature of GP, our model is capable of self-assessment. Large-scale experiments demonstrate that our method outperforms state-of-the-art techniques~\cite{Guillaumin2014ijcv,Guillaumin2012cvpr} for populating ImageNet with bounding-boxes and segmentations, as well as a strong MKL-SVM baseline defined on the same features.
\parshrinky
{ \paragraph{Acknowledgements}
This work was supported by ERC VisCul starting grant. A. Vezhnevets is also supported by SNSF fellowship PBEZP-2142889.}

\parskiny
{\footnotesize
\bibliographystyle{ieeetr}
\bibliography{./shortstrings,./calvin,./vggroup}
}
\end{document}